%% file: main.tex
\documentclass[lettersize,journal]{IEEEtran}
\usepackage{amsmath,amsfonts}
\usepackage{algorithmic}
\usepackage{array}
\usepackage[caption=false,font=normalsize,labelfont=sf,textfont=sf]{subfig}
\usepackage{textcomp}
\usepackage{stfloats}
\usepackage{url}
\usepackage{verbatim}
\usepackage{graphicx}
\usepackage{booktabs}
\usepackage{algorithm}
\usepackage{algorithmic}
\usepackage[switch]{lineno}
\usepackage{bbding}
\usepackage[cmyk]{xcolor}
\usepackage{stfloats}
\usepackage{amsfonts,amssymb}
\usepackage{colortbl}
\usepackage{hyperref}
\UseRawInputEncoding
\def\BibTeX{{\rm B\kern-.05em{\sc i\kern-.025em b}\kern-.08em
    T\kern-.1667em\lower.7ex\hbox{E}\kern-.125emX}}
\usepackage{balance}
\begin{document}
\title{GEOcc: Geometrically Enhanced 3D Occupancy Network with Implicit-Explicit Depth Fusion and Contextual Self-Supervision}
\author{Xin Tan, Wenbin Wu, Zhiwei Zhang, Chaojie Fan, Yong Peng, Zhizhong Zhang, Yuan Xie, Lizhuang Ma
\thanks{Manuscript received xxx; revised xxxx; accepted xxxx. Date of publication
xxx; date of current version xxx. This work was supported in part by the National Natural Science Foundation of China (62302167, U23A20343, 62222602, 62176092, and 62476090); in part by Shanghai Sailing Program (23YF1410500); Natural Science Foundation of Shanghai (23ZR1420400); in part by Chenguang Program of Shanghai Education Development Foundation and Shanghai Municipal Education Commission (23CGA34), in part by CCF-Tencent RAGR20240122. (Xin Tan and Wenbin Wu are co-first authors. Zhizhong Zhang is the corresponding author.) 

Xin Tan, Wenbin Wu, Zhizhong Zhang, Yuan Xie and Lizhuang Ma are with the School of Computer Science and Technology, East China Normal University, Shanghai 200062, China (e-mails:  xtan@cs.ecnu.edu.cn; 51255901049@stu.ecnu.edu.cn; 
zzzhang@cs.ecnu.edu.cn; xieyuan8589@foxmail.com; 
lzma@cs.ecnu.edu.cn).

Zhiwei Zhang is with the School of Electronics, Information and Electrical Engineering, Shanghai Jiao Tong University, Shanghai, China (e-mail: zhangzw12139@sjtu.edu.cn).

Chaojie Fan and Yong Peng are with the School of Transportation and Traffic Engineering, Central South University, China (e-mails:  fcjgszx@csu.edu.cn; 
yong\_peng@csu.edu.cn).
}
}

\markboth{Journal of \LaTeX\ Class Files,~Vol.~18, No.~9, September~2020}%
{How to Use the IEEEtran \LaTeX \ Templates}

\maketitle

\begin{abstract}

 3D occupancy perception holds a pivotal role in recent vision-centric autonomous driving systems by converting surround-view images into integrated geometric and semantic representations within dense 3D grids. Nevertheless, current models still encounter two main challenges: modeling depth accurately in the 2D-3D view transformation stage, and overcoming the lack of generalizability issues due to sparse LiDAR supervision. 
 To address these issues, this paper presents GEOcc, a Geometric-Enhanced Occupancy network tailored for vision-only surround-view perception. Our approach is three-fold: 1) Integration of explicit lift-based depth prediction and implicit projection-based transformers for depth modeling, enhancing the density and robustness of view transformation. 
 2) Utilization of mask-based encoder-decoder architecture for fine-grained semantic predictions; 3) Adoption of context-aware self-training loss functions in the pertaining stage to complement LiDAR supervision, involving the re-rendering of 2D depth maps from 3D occupancy features and leveraging image reconstruction loss to obtain denser depth supervision besides sparse LiDAR ground-truths. 
 Our approach achieves \textbf{State-of-the-Art} performance on the Occ3D-nuScenes dataset with \textbf{the least image resolution needed} and \textbf{the most weightless image backbone} compared with current models, marking an improvement of 3.3\% due to our proposed contributions. Comprehensive experimentation also demonstrates the consistent superiority of our method over baselines and alternative approaches.
 Our code is available at \href{https://github.com/world-executed/GEOcc.git}{https://github.com/world-executed/GEOcc.git}
\end{abstract}


\input{introduction}

\input{related_work}
\input{method}

\input{exp}

\input{conclusion}

\bibliographystyle{IEEEtran}
\bibliography{ref}

\begin{IEEEbiography}[{\includegraphics[width=1in,height=1.25in,clip,keepaspectratio]{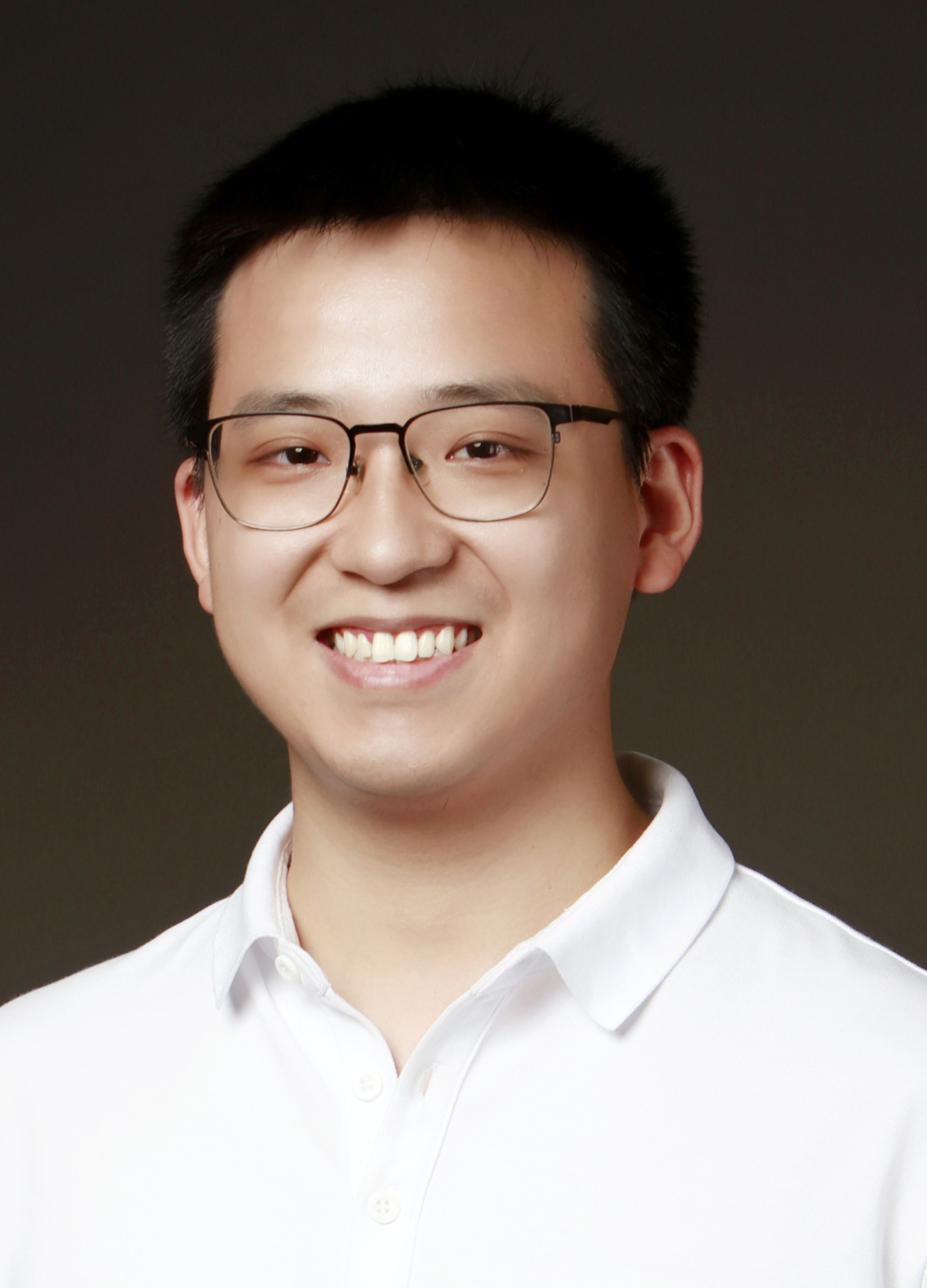}}]{Xin Tan} received dual Ph.D. degrees in Computer Science from Shanghai Jiao Tong University and City University of Hong Kong. He received his B.Eng. degree in Automation from Chongqing University, China in 2017.
He is currently an Associate Research Professor at the  School of Computer  Science and Technology, East China Normal University, China.
His research interests lie in computer vision and deep learning. He serves as a program committee member/reviewer for CVPR, ICCV, ECCV, AAAI, IJCAI, IEEE TPAMI, TIP and IJCV. He also serves as the associate editor for Pattern Recognition and Visual Computer. 
\end{IEEEbiography}

\begin{IEEEbiography}[{\includegraphics[width=1in,height=1.25in,clip,keepaspectratio]{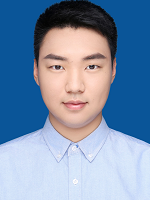}}]{Wenbin Wu} received the B.E. degree from Donghua University, Shanghai, China, in 2022. He is currently pursuing the M.E. degree with the School of Computer Science and Technology, East China Normal University. His current research interests include autonomous driving and computer vision.
\end{IEEEbiography}

\begin{IEEEbiography}[{\includegraphics[width=1in,height=1.25in,clip,keepaspectratio]{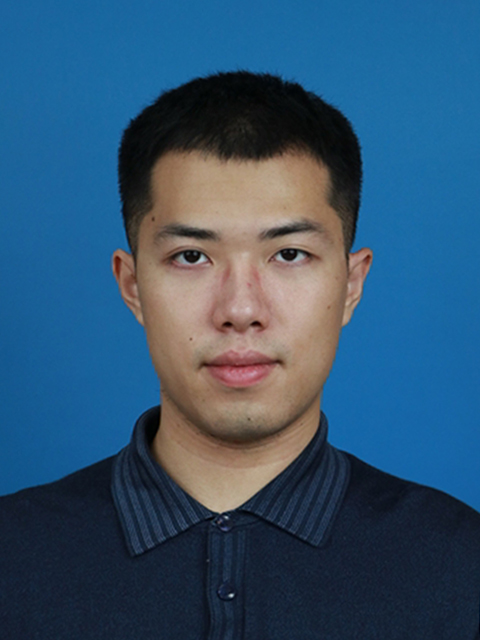}}]{Zhiwei Zhang} received the B.E. degree from Shandong University, Qingdao, China, in 2021. He is presently enrolled as a PhD candidate in Computer Science at the School of Electronics, Information and Electrical Engineering (SEIEE) of Shanghai Jiao Tong University. His research primarily explores multi-modal learning, vision-based perception in autonomous driving, and computer vision.
\end{IEEEbiography}

\begin{IEEEbiography}[{\includegraphics[width=1in,height=1.25in,clip,keepaspectratio]{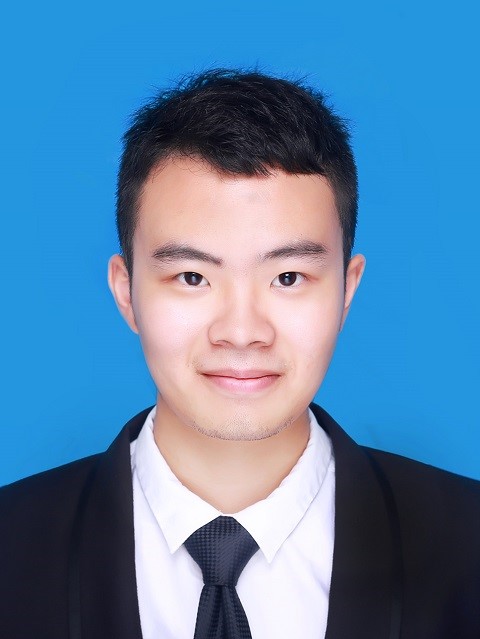}}]{Chaojie Fan} received his Ph.D. degree jointly from Central South University, China, and City University of Hong Kong, Hong Kong, China, in 2024. He is currently an Assistant Professor at Central South University. His research interests focus on brain–computer interaction and its application in intelligence transportation systems.
\end{IEEEbiography}

\begin{IEEEbiography}[{\includegraphics[width=1in,height=1.25in,clip,keepaspectratio]{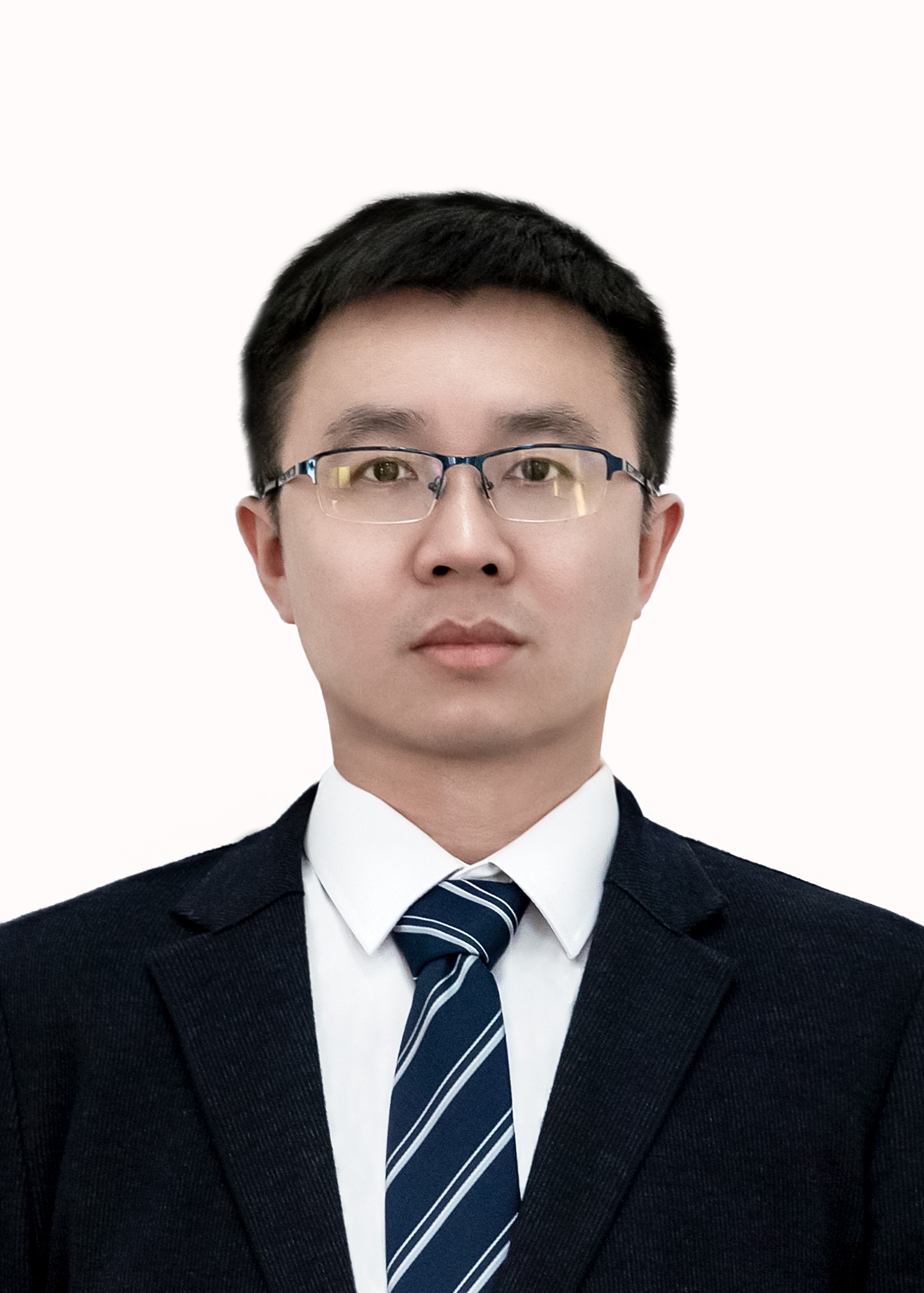}}]{Yong Peng} received the Ph.D. degree from the Strasbourg of University, France, in 2012. He is currently a Professor with the School of Transportation and Traffic Engineering, Central South University, China. His research interests include intelligence transportation, computer vision, and pattern recognition.
\end{IEEEbiography}

\begin{IEEEbiography}[{\includegraphics[width=1in,height=1.25in,clip,keepaspectratio]{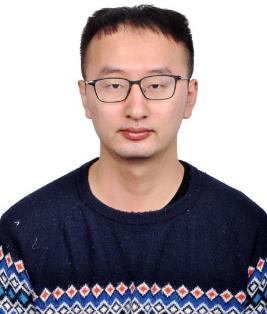}}]{Zhizhong Zhang} received the Ph.D. degree in pattern recognition and intelligent systems from the Institute of Automation, Chinese Academy of Sciences (CAS), in 2020. He is currently an Associate Professor with the  School of Computer  Science and Technology, East China Normal University.  His research interests include image processing, computer vision, machine learning, and pattern recognition.  
\end{IEEEbiography}

\begin{IEEEbiography}[{\includegraphics[width=1in,height=1.25in,clip,keepaspectratio]{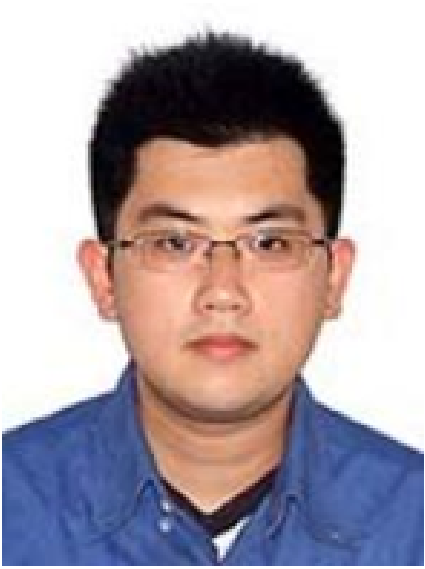}}]{Yuan Xie} received the PhD degree in
Pattern Recognition and Intelligent Systems from
the Institute of Automation, Chinese Academy of
Sciences (CAS), in 2013. He is currently a full
professor with the School of Computer Science and
Technology, East China Normal University, Shanghai,
China. His research interests include image
processing, computer vision, machine learning and
pattern recognition. He has published around 85 papers
in major international journals and conferences
including the IJCV, IEEE TPAMI, TIP, TNNLS,
TCYB, and NIPS, ICML, CVPR, ECCV, ICCV, etc.
He also has served as a reviewer for more than 15 journals and conferences.
Dr. Xie received the National Science Fund for Excellent Young Scholars 2022.

\end{IEEEbiography}

\begin{IEEEbiography}[{\includegraphics[width=1in,height=1.25in,clip,keepaspectratio]{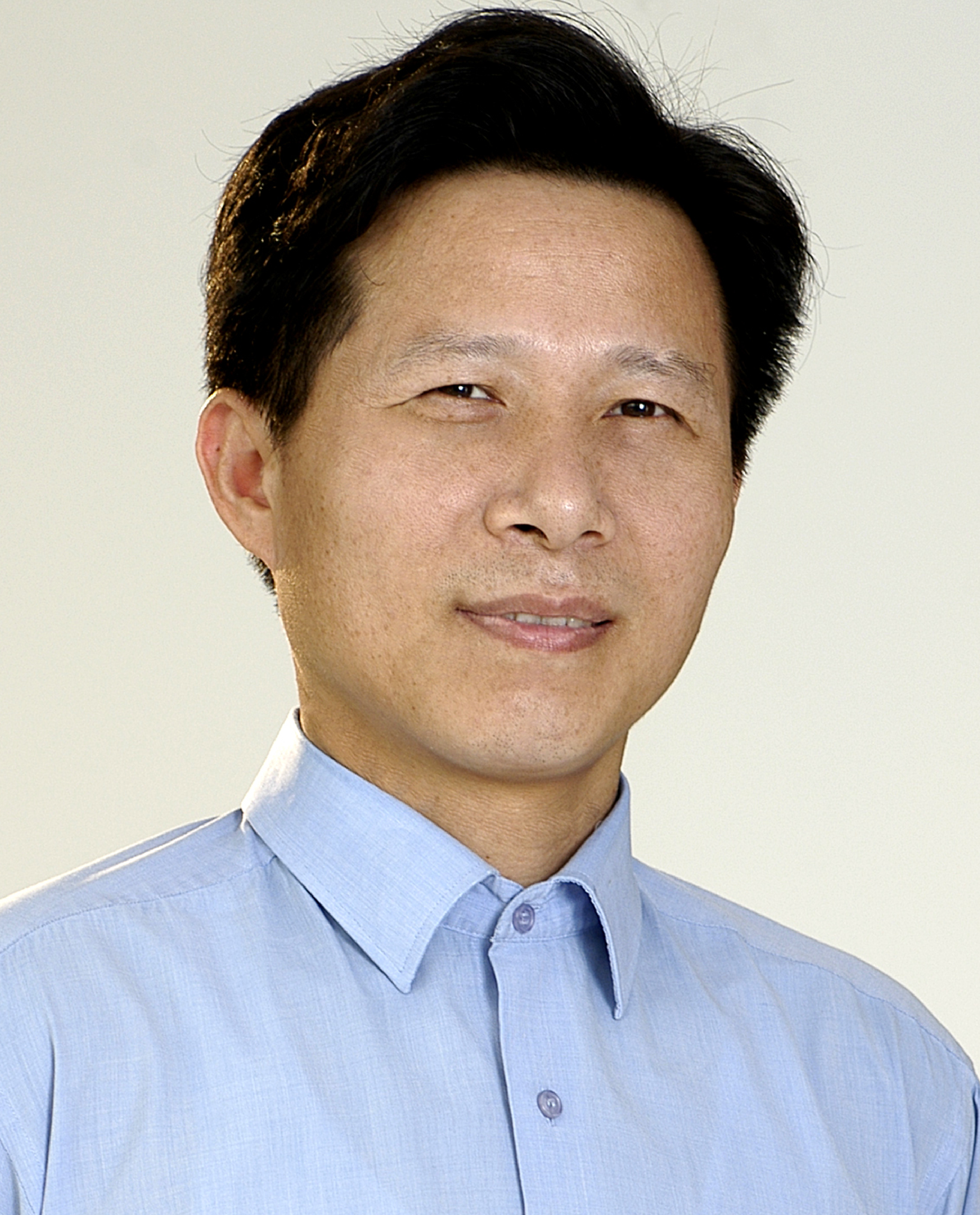}}]{Lizhuang Ma}
received his B.S. and Ph.D. degrees from Zhejiang University, China in 1985 and 1991, respectively. He is now a Distinguished Professor, at the Department of Computer Science and Engineering, Shanghai Jiao Tong University, China and the School of Computer Science and Technology, East China Normal University, China. 
His research interests include computer vision, computer aided geometric design, computer graphics, scientific data visualization, computer animation, digital media technology, and theory and applications for computer graphics, CAD/CAM.
\end{IEEEbiography}

\end{document}

%% file: introduction.tex
\begin{figure}[!htbp]
 \centering
 \includegraphics[width=\columnwidth]{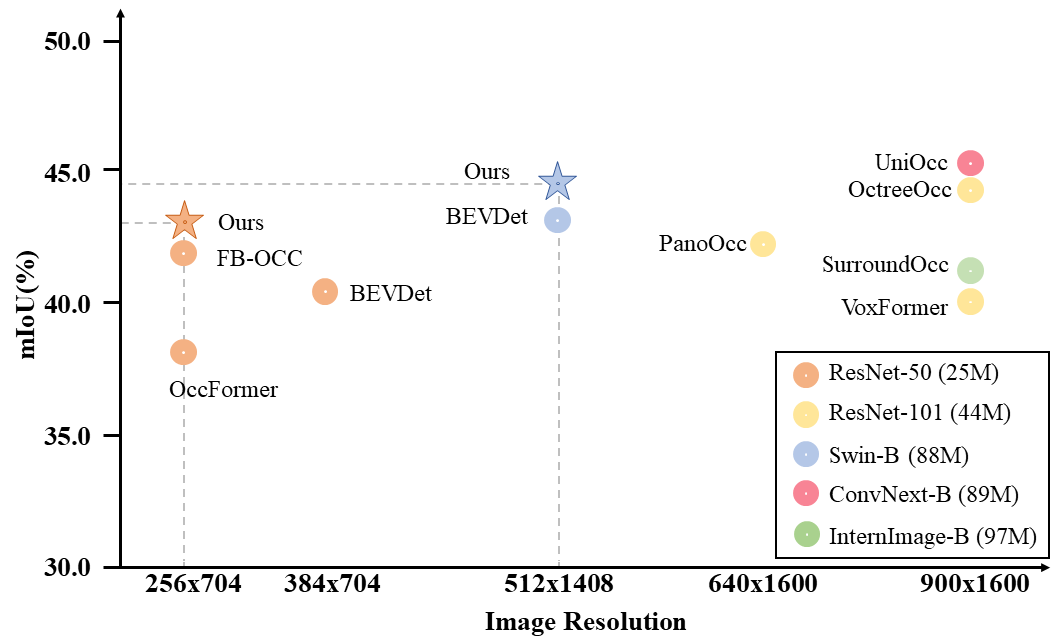} 
 \setlength{\abovecaptionskip}{-0.3cm}
 \setlength{\belowcaptionskip}{-0.5cm}
 \caption{
    Performance comparison with other state-of-the-art methods. Different colors represent various image backbones, from the smallest ResNet-50 to the largest InternImage-B in the experiment settings of existing approaches. Our GEOcc network gets leading performance among others using lighter image backbones (ResNet-50 and SwinTransformer-B) and lower image resolutions. 
 }
 \label{fig:sec1_performance}
\end{figure}

\section{Introduction}

3D vision for autonomous driving \cite{su2024co,wang2022channel,wang2020low,sun2024multi,zhang2021dpsnet,liu2024service,cvm2025} has undergone a significant shift from LiDAR-based multi-modal fusion to image-based vision-centric perception in recent years, primarily driven by the potential to reduce reliance on costly LiDAR sensors.  The occupancy network has emerged as a cornerstone in vision-centric methodologies by transforming surround-view perspective images into a grid-based representation of the nearby environment. This includes both the top-down Bird-Eye-View (BEV) occupancy and voxel-like 3D occupancy formats. Compared to BEV occupancy, 3D occupancy enables a more fine-grained description of height information for various objects, \textit{e.g.}, poles, traffic signs, and bridges. Consequently, high-fidelity 3D representations depend on precise depth prediction and fine-grained semantic features from 2D images.

Despite the increasing research focus on the design of 3D occupancy networks, a common challenge encountered is training a robust 2D-to-3D view transformation network, which essentially impacts the quality and accuracy of downstream occupancy prediction. This challenge stems from two primary factors: 1) The 2D-to-3D transformation relies on accurate 3D position estimation of various objects at multiple scales. However, 2D-to-3D depth prediction is inherently an ill-posed problem; 2)  The occupancy ground-truths in most open-source dataset \cite{tian2023occ3d,wang2023openoccupancy} are typically generated from sparse LiDAR points, leading to sparsity issues that restricts the generalizability of dense occupancy features. Specifically, numerous occupancy grids that coincide with rich image features are masked out during training. Furthermore, relying solely on LiDAR supervision may result in the overfitting of existing occupancy models.

For robust view transformation, explicit depth modeling (EDM) and implicit depth modeling (IDM) are separately adopted by mainstream occupancy networks. EDM employs a designated prediction network to extrapolate the depth probability distribution from pixels in images to a 3D space, represented by LSS \cite{Philion_2020lss}, BEVDet \cite{huang2021bevdet} and OpenOccupancy \cite{wang2023openoccupancy}. The merit of EDM lies in its compartmentalized depth prediction process, facilitating precise depth estimation; however, the sparsity of LiDAR points limits the supervision of pixel depth.  Conversely, IDM leverages cross-attention and self-attention mechanisms to directly transition image features to occupancy features, with implementations such as BEVFormer \cite{huang2022bevdet4d}, TPVFormer \cite{huang2023tri}, among others. IDM's strength is in its enhanced differentiability and the generalizability of occupancy features, owing to its end-to-end transformation approach, though it can result in depth confusion. 
By combining both methods, we can not only address the distribution uncertainty in explicit depth estimation but also mitigate feature redundancy in implicit depth modeling.

For sparse LiDAR supervision, the mainstream approach is to densify LiDAR points to cover more occupancy grids \cite{wang2023openoccupancy,tian2023occ3d} by mesh reconstruction, temporal accumulation, etc. However, inspired by current self-supervised depth estimation methods like VFDepth \cite{kim2022self}, we hypothesize that image-based self-supervision may alleviate sparse LiDAR issues in another novel approach. Given that occupancy features are derived from image features,  self-supervision applied to the 2D perspective view can effectively propagate the supervisory signal back to occupancy features. 
Through this self-supervised learning approach, the network can acquire rich geometric structural priors. Moreover, this prior information is denser than what is obtained through training directly on manually annotated ground truth.
Meanwhile, self-supervision can exploit information from multiple cameras and their temporal correlations, which largely extends the scope of  LiDAR supervision signals.

To tackle the challenges of view transformation and sparse LiDAR supervision, we introduce GEOcc, a Geometric-Enhanced occupancy network tailored for vision-only surround-view perception. Our approach is three-fold: 1) An explicit lift-based and implicit projection-based depth prediction framework to improve the occupancy feature density and robustness. To be specific, we propose Implicit Depth Modeling (IDM),  Explicit Depth Modeling (EDM), and an efficient depth fusion module. 
2) A mask-based encoder-decoder structure for detailed semantic predicting, utilizing a transformer approach to refine occupancy features and generate semantic predictions, as opposed to the conventional convolution-based decoders \cite{wang2023openoccupancy}.
3)  A context-aware self-training loss mechanism that involves the re-rendering of 2D depth maps from 3D occupancy features and image reconstruction loss besides sparse LiDAR supervision. We adopt a label-free self-supervision strategy for fair comparison, which encompasses spatial, temporal, and spatial-temporal camera supervision applicable during the pre-training phase.

To summarize, our contributions are as follows:

\begin{itemize}
\item We propose a novel hybrid depth modeling framework for better 2D-to-3D view transformation, including Explicit Depth Modeling (EDM) and implicit Depth Modeling (IDM), integrated via a lightweight fusion module. This hybrid modeling process enhances the robustness and generalizability of 3D occupancy features.

\item We have pioneered the incorporation of 2D contextual self-supervision into the training of our 3D occupancy network. This method utilizes spatial, temporal, and spatial-temporal image reconstruction loss among rendered depth maps to mitigate the limitations posed by LiDAR sparsity.

\item We have successfully integrated these innovations into a mask-based transformer encoder-decoder architecture, enabling end-to-end training. Our methodology has demonstrated a \textbf{State-Of-The-Art} performance (44.7\% mIoU) on the Occ3D-nuScenes dataset with \textbf{the least image resolution needed} and \textbf{the most weightless image backbone} compared to prevailing methods, and an \textbf{3.3\%} mIoU improvement due to our proposed contributions.
\end{itemize}

%% file: related_work.tex
\section{Related Work}
\subsection{Explicit-Implicit View Transformation.}
Vision-based 3D perception has been widely studied in recent years \cite{ma2022vision} due to its advantages in cost-effectiveness and straightforward data collection. Initial efforts focused on transforming 2D image features into Bird's Eye View (BEV) planes for improved detection and planning capabilities, utilizing depth cues for precise 3D localization. The pioneering research LSS \cite{Philion_2020lss} and the following BEV-based detection methods \cite{huang2021bevdet,huang2022bevdet4d} explicitly predict the depth distribution of each pixel to project 2D features to 3D features. 
 These approaches often struggle with uncertainties in depth estimation, which can introduce errors during the feature transformation process. Additionally, the generated features are typically highly sparse, limiting their effectiveness in downstream networks.
Subsequent research introduced implicit methods, evolving from BEV to full 3D occupancy perception \cite{li2022bevformer,huang2023tri,wei2023surroundocc,zhang2023occformer,pan2023renderocc,li2023voxformer}, employing attention mechanisms to learn view transformations. 
However, this methods may result in interactions between multiple voxels along the same ray and the same pixel, leading to feature redundancy and duplication. In contrast, our approach integrates explicit and implicit depth modeling and leverages 3D convolutions for effective feature fusion and compression. This combination significantly enhances feature robustness while mitigating redundancy.
Despite these advancements, few have effectively integrated both implicit and explicit depth approaches for 3D occupancy under vision-only conditions. For example, FB-OCC \cite{li2023fb} attempts to enhance explicit depth predictions using BEVFormer-based backward projection, yet its implicit BEV modeling omits vertical dimension details. Our work, GEOcc, aims to effectively merge the strengths of both depth paradigms. Unlike FB-OCC, GEOcc leverages a combined 3D occupancy representation through concerted implicit and explicit modules, complemented by an innovative occupancy compressor and a multi-scale Mask-based Encoder-Decoder structure to further refine occupancy features.

\subsection{3D Occupancy Prediction}
The task of 3D occupancy prediction has garnered increasing attention in recent years, emerging from semantic scene reconstruction (SSC) tasks.
Some previous work \cite{trans4,trans5,trans7,trans8}has explored the use of Occupancy grid maps to sense the environment for autonomous driving. However, due to the lack of semantic categories, simple 0-1 categorization does not provide a rich understanding of the environment.
\cite{trans6}introduced semantic segmentation and attempts to utilize raw radar data as input.\cite{trans2}
bring bayesian learning to occupancy tasks.
3D Occupancy Prediction area can be categorized into three primary research streams: LiDAR-based occupancy prediction (related to completion tasks) \cite{khurana2022differentiable,mahjourian2022occupancy,khurana2023point,agro2023implicit}, camera-only occupancy predictions \cite{zhang2023occformer,li2023fb,pan2023uniocc,huang2023selfocc,lu2023octreeocc,pan2023renderocc}, and integrated LiDAR-Camera Occupancy predictions \cite{wang2023openoccupancy,zhang2023radocc}. LiDAR can serve as an informative prompt to substantially improve the accuracy of current occupancy prediction results enabling the forecasting of 4D occupancy dynamics over an imminent time frame \cite{khurana2022differentiable,khurana2023point}. However, camera-only occupancy predictions are more challenging for accuracy improvement because of the robustness and generalizability bottleneck in 2D-to-3D view transformation without LiDAR input. COTR\cite{ma2023cotr} focuses on compact representation, but lacks of rich 2D supervisory information on view transformations. To this end, our GEOcc focuses on vision-only occupancy predictions and attempts to geometrically enhance the view transformation process. 
Recent works, such as SparseOcc\cite{tang2024sparseocc} and FlashOcc\cite{yu2023flashocc}, have made progress in memory-efficiency. However, compared to GEOcc, they lack dense self-supervised pretraining and robust 3D feature representation. This is the key reason why GEOcc achieves a higher mIoU.

\subsection{Volume Rendering for Self-Supervision}
Recent occupancy models, such as RenderOcc \cite{pan2023renderocc}, UniOcc \cite{pan2023uniocc}, and Self-Occ \cite{huang2023selfocc} begin to use volume rendering\cite{mildenhall2021nerf,trans1} to get supervision form images\cite{trans3}. However, their performance is generally less competitive when compared with leading methodologies. We hypothesize that the simultaneous rendering of depth and semantic maps for self-supervision significantly complicates the convergence of 3D occupancy models. As geometric reconstruction is fundamental to accurate occupancy modeling we choose to utilize pure geometric pretraining through volume rendering techniques. our method distinguishes itself from the aforementioned studies by introducing a tripartite contextual self-supervision loss: spatial, temporal, and spatial-temporal. These three sets of losses provide more dense supervision.

%% file: method.tex
\begin{figure*}[!htbp]
 \centering
 \includegraphics[width=1.0\textwidth]{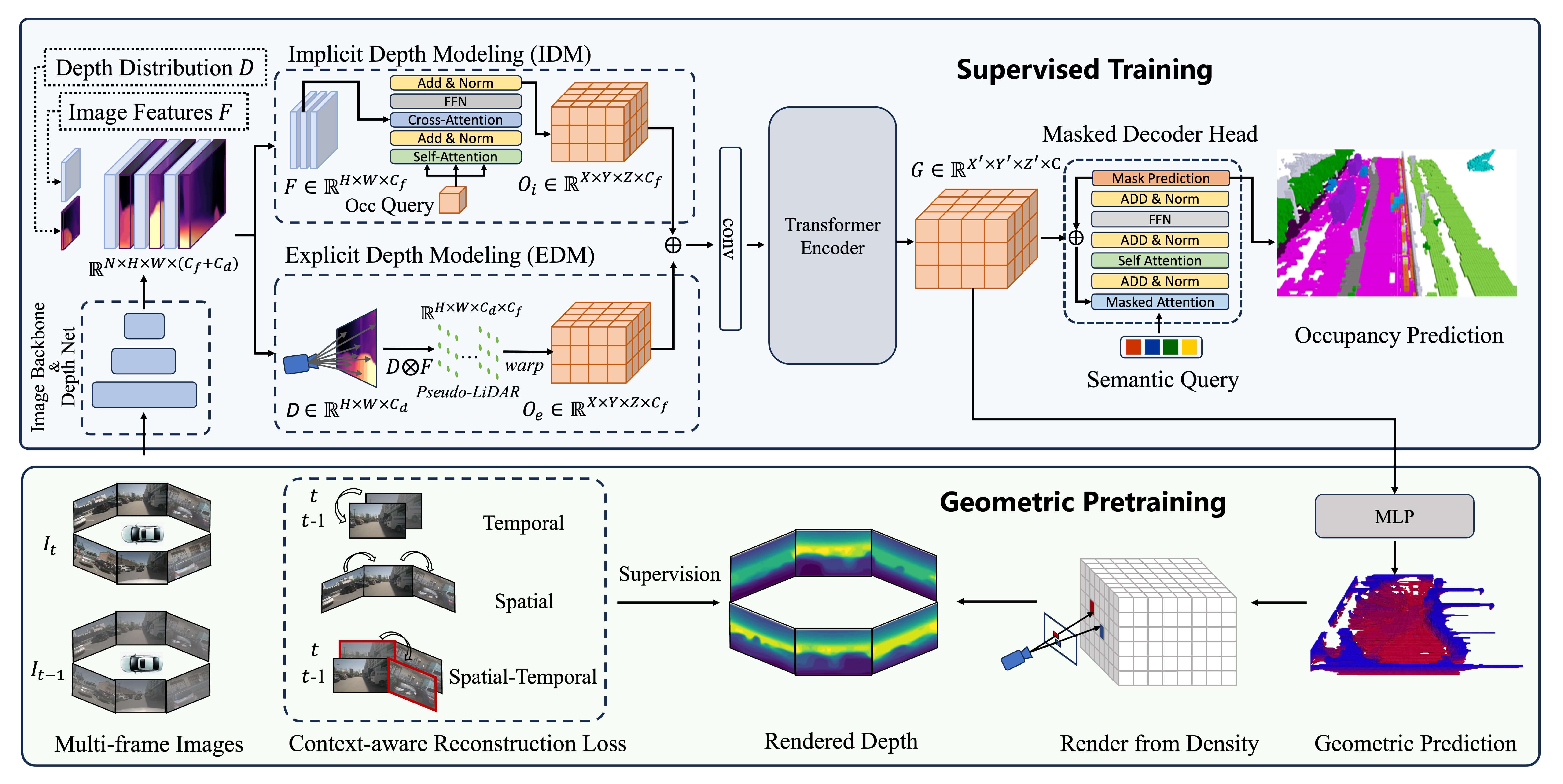} 
 \setlength{\abovecaptionskip}{-0.3cm}
 \setlength{\belowcaptionskip}{-0.5cm}
 \caption{
 The GEOcc network architecture processes surround-view images from multiple timestamps, with an image backbone to extract multi-scale image features $\mathbf{F}$ and depth distribution features $\mathbf{D}$. The Implicit Depth Modeling (IDM) performs self-attention and projection-based deformable cross-attention to generate implicit occupancy features $\mathbf{O}_\textbf{I}$, while the Explicit Depth Modeling (EDM) performs cross-product between lifted depth prediction and image features to create grid-like pseudo-LiDAR points, which are then warped into explicit occupancy feature $\mathbf{O}_\textbf{E}$. Occupancy features are subsequently fused and fed into a transformer encoder and a Masked Decoder Head at different resolutions to yield final predictions. In the geometric pretraining period, an extra MLP layer predicts geometric density, which is subsequently rendered into surround-view perspective depth maps. Finally, geometric pretraining is accomplished through the application of three types of Context-Aware Self-Training (CAST) losses. Better zoomed in.
 }
 \label{fig:sec2_mainmodel}
\end{figure*}

\section{Method}
 
\subsection{Problem Setup}
The 3D occupancy prediction task aims to infer whether each voxel in the 3D space is occupied by objects, along with its semantic labels, given surround-view 2D camera images as inputs. We define the input images as $\mathbf{I}_i^t\in\mathbb{R}^{H \times W \times 3}$ , where $i\in\{1,2,\dots,N\}$ denotes the $i$-$th$ of $N$ cameras, and $t\in\{T,T-1,...,T-\tau\}$ denotes the current timestamp at $T$ with historical $\tau$ frames. $H$ and $W$ represent the height and width of images. Then, 3D occupancy prediction can be formulated as follows:
\begin{equation}
    \mathbf{O}=\mathbb{V}(\mathbb{F}_{\text{img}}(\mathbf{I}_i^t)), \mathbf{G}=\mathbb{F}_{\text{enc}}(\mathbf{O}), \mathbf{P}=\mathbb{F}_{\text{dec}}(\mathbf{G}),
\end{equation}
where $\mathbb{F}_{\text{img}}$ is an image backbone network, $\mathbb{V}$ is a 2D-to-3D view transformation function, $\mathbb{F}_{\text{enc}}$ and $\mathbb{F}_{\text{dec}}$ are voxel encoder and decoder, respectively. $\mathbf{O}$, $\mathbf{G}$, and $\mathbf{P}$ correspond to the initial occupancy features, the refined occupancy features, and the final classification predictions. 
Typically, we perform a $K+1$ classification for each voxel. It includes $K$ semantic labels which are inherently considered occupied and one additional {\it ``free''} label indicating an unoccupied state.

\subsection{Network Structure}

Our framework is shown in Fig. \ref{fig:sec2_mainmodel}, which contains three key components. First, we use both explicit and implicit 2D-to-3D view transformation to get 3D occupancy features from 2D image features. Then, the mask-based encoder-decoder structure is used to obtain comprehensive occupancy representation and per-voxel semantic labels. Finally, to enhance the geometric perception of our occupancy network, we introduce self-supervised geometric pretraining to initialize network weights.

\noindent{\bfseries 2D-to-3D View Transformation.}
In vision-only occupancy tasks, 3D occupancy features are often inaccurate due to the absence of explicit depth information or overfitted owing to the lack of implicit correlations. To improve the quality of these features, we implement a combination method of explicit and implicit depth modeling. For explicit depth modeling, an auxiliary depth net is used to predict the depth distribution $\mathbf{D} \in \mathbb{R}^{H \times W \times C_d}$ for each pixel, where $C_d$ represents the number of depth bins. With image features $ \mathbf{F} \in \mathbb{R}^{H \times W \times C_f} $ extracted by the image backbone, where $C_f$ denotes the number of image channels, we perform broadcastable outer product $ \mathbf{D} \otimes \mathbf{F}$ to lift pixels into grid-like pseudo-LiDAR points $\mathbb{R}^{H \times W \times C_d \times C_f}$ in camera coordinate. Then we transform pseudo points to world coordinates, warp them into new voxel grids of fixed resolution $X \times Y \times Z$ based on their 3D positions, and finally conduct 3D voxel-pooling to generate the explicit occupancy feature $\mathbf{O}_\text{E} \in\mathbb{R}^{\times X \times Y \times Z \times C_f}$.

For implicit depth modeling, we predefine learnable occupancy queries $\mathbf{Q}\in\mathbb{R}^{X \times Y \times Z \times C_f}$, to which self-attention and deformable cross-attention \cite{zhu2020deformable} are applied. To be specific, each 3D point in the query voxel is projected onto the image plane to sample features for associated keys and values. Then, a deformable cross-attention layer is adopted to implicitly transform 2D features to 3D occupancy features $\mathbf{O}_\text{I}\in\mathbb{R}^{X \times Y \times Z \times C_f}$. The final occupancy features are concatenations of $\mathbf{O}_\text{E}$ and $\mathbf{O}_\text{I}$. The whole 2D-to-3D view transformation can be denoted as follows:
\begin{equation}
\begin{array}{rl}
    \mathbf{O}_\text{E} =& \operatorname{Pooling}(\operatorname{Warp}(\mathbf{E}^{-1}(\mathbf{D} \otimes \mathbf{F}))) \\
    \mathbf{O}_\text{I} =& \operatorname{DeformAttention}(\mathbf{Q}, \mathbf{F}, \mathbf{E}, \mathbf{K}) \\
    \mathbf{O} =& \mathbf{O}_\text{E} \oplus \mathbf{O}_\text{I},
\end{array}
\end{equation}
\noindent where $\mathbf{E}$ and $\mathbf{K}$ are camera extrinsic and intrinsic parameters and $\oplus$ means concatenation in feature dimension.
Some traditional methods, such as BEVDet\cite{huang2021bevdet}, BEVStereo\cite{li2023bevstereo}, and OccFormer\cite{zhang2023occformer}, rely solely on explicit depth estimation to lift 2D image features into 3D space. These approaches often suffer from uncertainties in depth estimation, leading to errors during the feature transformation process. Moreover, the resulting features tend to be highly sparse. Other methods, such as SurroundOcc\cite{wei2023surroundocc} and BEVFormer\cite{li2022bevformer}, employ cross-attention between 3D and 2D for view transformation. However, this can result in interactions between multiple voxels along the same ray and the same pixel, causing feature redundancy and duplication. In contrast, our method combines both explicit and implicit depth modeling, and uses 3D convolutions for feature fusion and compression, thereby enhancing feature robustness.

\noindent{\bfseries Occupancy Feature Compression Layer.}
The occupancy features derived from explicit depth modeling usually have scene sparsity issues due to discrete depth bins, while implicit depth modeling tends to induce feature redundancy as multiple voxels are often projected onto the same image region. To mitigate these two problems, we use a compression layer to intensify occupancy features after concatenation. This layer, essentially a convolutional layer with a stride of 2, reduces the occupancy resolution from $(X, Y, Z)$ to $(X/2, Y/2, Z/2)$. The concatenation of EDM and IDM mitigates the sparsity issue of EDM and the redundancy issue of IDM. The convolution layer further reduces the number of feature parameters, resulting in compact occupancy features.

\noindent{\bfseries Mask-based Transformer Encoder-Decoder Structure.}
The occupancy features $\mathbf{O}$ obtained through implicit-explicit depth modeling require more effective integration and comprehension. Drawing inspirations from OccFormer \cite{zhang2023occformer}, we use a transformer encoder, denoted as $\mathbb{F}_\text{enc}$, to generate multi-scale occupancy features. Our encoder layer contains a window-divided self-attention and a downsampling convolution, which can be formulated as,
\begin{equation}
    \operatorname{Attention}(\mathbf{Q}, \mathbf{K}, \mathbf{V})=\operatorname{SoftMax}\left(\mathbf{Q} \mathbf{K}^{T} / \sqrt{d}+\mathbf{B}\right) \mathbf{V}.
\end{equation}
In this equation, $\mathbf{Q},\mathbf{K}, \mathbf{V}\in\mathbb{R}^{N_\text{window}\times d}$ represent the query, key, and value matrices, respectively; $\mathbf{B}\in\mathbb{R}^{N_\text{window}\times N_\text{window}}$ is the bias matrix; $N_\text{window}$ denotes the number of patches in each window; and $d$ indicates the dimension. After the window-based attention, the downsampling convolution yields multi-scale occupancy features denoted by $\mathbf{G}_l$ for scales $l\in\{1\dots L\}$. We then utilize multi-scale deformable attention, as proposed in \cite{cheng2021mask2former}, to amalgamate these features into the aggregated occupancy features $\mathbf{G}$.

For decoder, we apply Mask2Former \cite{cheng2021mask2former} decoder head on $\mathbf{G}_l$, where the masked attention operation can be formulated as,
\begin{equation}
\mathbf{X}_{l+1}=\operatorname{SoftMax}\left(f_q\left(\mathbf{X}_{l} \right) f_k\left(\mathbf{G}_{l} \right)^{\mathrm{T}}+\mathcal{M}_{l}\right) 
f_v\left(\mathbf{G}_{l} \right)+\mathbf{X}_{l},
\label{eq: mask2former}
\end{equation}
where $\mathbf{X}_{l} \in \mathbb{R}^{N_q \times C}$ denotes $N_q$ semantic queries with $C$-dimension in $l$-th level and $\mathcal{M}_{l}$ denotes the binary mask predicted from previous layer. 
The linear transformations $f_q(\cdot)$, $f_k(\cdot)$, and $f_v(\cdot)$ remap semantic queries $\mathbf{X}_l$ and occupancy features $\mathbf{G}_l$ to new feature spaces as queries, keys, and values, respectively. Following the masked attention step, a Feed-Forward Network (FFN) layer is employed for feature projection. We use addition and normalization to construct a residual connection. Ultimately, the final query features $\mathbf{X}_l$ are processed by MLP to predict the semantic class $c_i$, and the corresponding binary 3D mask $\mathcal{M}_i$ for $c_i$ is generated by Eq. \ref{eq: mask2former}.
The mask-based semantic decoder demonstrates strong semantic segmentation capabilities. In works such as MaskFormer\cite{cheng2021per} and the subsequent Mask2Former\cite{cheng2021mask2former}, the semantic segmentation task is treated as a binary mask prediction for each category. Given its success in 2D image tasks, we introduce this approach to the 3D occupancy prediction task. We replace the original 2D attention mechanism with 3D attention and output 3D masks. Additionally, during the 3D feature encoding stage, we substitute the baseline 3D U-Net\cite{huang2021bevdet} structure with a Transformer-based model.

\subsection{Self-supervised Geometric Pretraining}

Previous research has demonstrated the significance of depth estimation in tasks related to occupancy prediction \cite{huang2021bevdet,li2023fb}. However, relying solely on sparse LiDAR data for depth and occupancy supervision often limits the generalizability of derived occupancy features. To address this issue, we propose the novel integration of Context-Aware Self-supervised Training (CAST) into the 3D occupancy network framework. The basic idea is to render surround view depth from occupancy features and use image reconstruction photometric loss to supervise the dense depth. The whole process is in a self-supervised manner without any need for additional 3D labels. We employ this loss function during the pretraining stage to enhance the geometric comprehension of our network.

\begin{center}
\begin{table*}[!htbp]
\caption{3D occupancy prediction performance comparison (ordered by mIoU) on the Occ3D-nuScenes dataset. Since some methods before the release of Occ3D-nuScenes only report LiDAR segmentation results instead of occupancy results, and some are BEV-based detection methods that are converted into occupancy segmentation tasks, we divide the reported results into official reports (no marks), our implementations (one $*$) and others' reports (two $**$). Especially for $**$, the report reference is from OctreeOcc. $\dagger$ indicates that the method adopts test-time augmentations, and \textit{CVPRW} denotes the \textit{CVPR} workshop papers.}
\resizebox{1.0\textwidth}{!}{
\setlength{\tabcolsep}{2.5mm}{
\begin{tabular}{l|c|c|c|c|c|c}
\toprule
Method      & Venue    & Image Backbone & Image Size & Visible mask  & 
IoU(\%) &mIoU(\%)  \\ 

\midrule
$*$ OccFormer \cite{zhang2023occformer}   & ICCV'2023 & ResNet-50 (25M)   &  256$\times$704     &
\XSolidBrush  & 70.1 & 37.4  \\
 * UniOcc\cite{pan2023uniocc} & CVPRW'2023& ResNet-50& 256$\times$704& \CheckmarkBold  & -&39.7\\
$*$ BEVDet4D \cite{huang2022bevdet4d}   & arXiv'2022 & ResNet-50      & 384$\times$704      &           \CheckmarkBold  &74.53 & 40.43     \\
$\dagger$ FBOcc \cite{li2023fb}      & CVPRW'2023  & ResNet-50      & 256$\times$704       & \CheckmarkBold    &-   & 42.14      \\
\rowcolor[HTML]{DFDFDF} \textbf{BEVDet4D + Ours} &  -    & ResNet-50      &  \textbf{256$\times$704}    &        \CheckmarkBold  & {\bfseries74.63}  &  {\bfseries 43.64}  \\

MonoScene \cite{cao2022monoscene}  & CVPR'2022  & ResNet-101 (44M)    & 92$\times$600      & \XSolidBrush   &-   & 6.06       \\
**BEVDet \cite{huang2021bevdet}     & arXiv'2021  & ResNet-101     & 256$\times$704    &
\XSolidBrush    &-   & 11.73      \\
**BEVFormer \cite{li2022bevformer}  & ECCV'2022  & ResNet-101      & 256$\times$704       & \XSolidBrush    &-    & 23.67      \\
**BEVStereo \cite{li2023bevstereo}  &  AAAI'2023   &  ResNet-101  & 256$\times$704   &
\XSolidBrush    &-   & 24.51  \\
**TPVFormer\cite{huang2023tri}  & CVPR'2023  & ResNet-101     & 900$\times$1600 &
\XSolidBrush   &-   & 28.34      \\
CTF-Occ \cite{tian2023occ3d}    & NIPS'2024 & ResNet-101     & 928$\times$600      & \XSolidBrush    &-  & 28.53      \\

VoxFormer \cite{li2023voxformer}   & CVPR'2023  & ResNet-101     & 900$\times$1600     & \CheckmarkBold   &-  & 40.71      \\

PanoOcc \cite{wang2023panoocc} & CVPR'2024 & ResNet-101 & 640$\times$1600 &
\CheckmarkBold  &-   & 42.13     \\
OctreeOcc \cite{lu2023octreeocc} & CVPR'2024 & ResNet-101  &  900$\times$1600   &   \CheckmarkBold  & - &   44.02 \\

RenderOcc   \cite{pan2023renderocc}   & ICRA'2024 & Swin-B (88M) & 512$\times$1408 & \XSolidBrush     &-   & 26.11      \\
*SurroundOcc \cite{wei2023surroundocc} & ICCV'2023  & InternImage-B (97M)  & 900$\times$1600    & \CheckmarkBold     &-   & 40.70      \\

*BEVDet4D \cite{huang2022bevdet4d}   & arXiv'2022 & Swin-B (88M) & 512$\times$1408     &       \CheckmarkBold &   72.47  & 43.29   \\
\rowcolor[HTML]{DFDFDF} \textbf{BEVDet4D + Ours} &  -   & Swin-B (88M) & \textbf{512$\times$1408}  &        \CheckmarkBold & {\bfseries72.64} &   {\bfseries44.67}      \\

\bottomrule
\end{tabular}}}
\setlength{\belowcaptionskip}{-0.4cm}

\label{tab: main results}
\end{table*}
\end{center}

\noindent{\bfseries Rendering Surround View Depth Map.}
To obtain depth maps for each viewpoint, we start with the occupancy features $\mathbf{G} \in \mathbb{R}^{X \times Y \times Z \times C'}$ procured from the transformer encoder $\mathbb{F}_{\text{enc}}$. $\mathbf{G}$ is firstly interpolated to full resolution, and an MLP layer is utilized to predict the per-voxel density $\boldsymbol{\sigma} \in \mathbb{R}^{X \times Y \times Z}$. For every pixel, a ray originating from the camera center $\mathbf{o}$ is drawn into the camera coordinates space in the direction $\mathbf{d}$. These rays $\mathbf{r}$ are described by the equation $\mathbf{r} = \mathbf{o} + t\mathbf{d}$, where $t$ controls the ray's length. Subsequently, we sample $S$ points at equal intervals along ray $\mathbf{r}$ to get the point set $\{p_i | i\in\{1,2,\dots,S\} \}$. Following NeRF-based volume rendering techniques \cite{mildenhall2021nerf,Khurana_2022}, the rendered depth value of a given ray $\mathbb{D}
(\mathbf{r})$ is determined by the following equations:
\begin{equation}
    \mathbb{T}\left(\mathbf{r}\right)=\exp (-\sum_{j=1}^{S-1} \boldsymbol{\sigma}\left(p_{j}\right) \boldsymbol{\delta}_{j}),
\end{equation}
\begin{equation}
    \mathbb{D}(\mathbf{r})=\sum_{i=1}^{S} \mathbb{T}\left(\mathbf{r}\right)\left(1-\exp \left(- \boldsymbol{\sigma} \left(p_{i}\right) \boldsymbol{\delta}_{i}\right)\right) p_{i}.
\end{equation}
$\mathbb{T}(\mathbf{r})$ is the accumulated transmittance of ray $\mathbf{r}$, $\boldsymbol{\delta}_{i} = p_{i+1}-p_{i}$ is the interval between two consecutive points, and $\boldsymbol{\sigma} \left(p_{i}\right)$ denotes the density at the $i$-th sample point. Such a process is the discrete approximation of the continuous neural rendering, transforming a 3D density field into 2D depth maps.

\noindent{\bfseries Context-Aware Self-Training Loss.}
Besides the sparse depth supervision form projected LiDAR, we use chronological surround-view images $\mathbf{I}_i^t$ to provide dense supervision for the self-training paradigm, wherein $i$ denotes the camera index and $t$ represents the timestamp. Given rendered depth maps $\mathbb{D}(\mathbf{r})$ and ego-car pose matrix $\mathbf{T}^{t \rightarrow t'}$, it becomes feasible to warp the reference image $\mathbf{I}_i^t$ into the reconstructed new image $\tilde{\mathbf{I}}_{i'}^{t'}$ as perceived from a neighboring $i'$-$th$ camera at time $t'$. Following VFDepth \cite{kim2022self}, there are three specific types of neighboring contexts: temporal $(\mathbf{I}_i^t \rightarrow \mathbf{I}_i^{t'})$, spatial $(\mathbf{I}_i^t \rightarrow \mathbf{I}_j^t)$ and spatial-temporal $(\mathbf{I}_i^t \rightarrow \mathbf{I}_j^{t'})$ correlations, particularly when $i'=j \ne i$. The warp function can be formulated as:
\begin{equation}
\begin{array}{c}
    \mathbf{M}_{\text{t}} = \mathbf{T}_i^{t \rightarrow t'} \\
    \mathbf{M}_{\text{sp}} = \mathbf{E}_j\mathbf{E}_i^{-1} \\
    \mathbf{M}_{\text{spt}} = \mathbf{E}_{j} \mathbf{E}_i^{-1}\mathbf{T}_i^{t \rightarrow t'},
\end{array}
\end{equation}

\begin{equation}
\begin{array}{cl}
    &\tilde{\mathbf{I}}_{i'}^{t'} = \mathbf{K}_{i'} \mathbf{M} \mathbb{D}(\mathbf{r}) \mathbf{K}_i ^{-1} \mathbf{I}_i^t,
\end{array}
\end{equation}

\noindent where $\mathbf{M}\in \{\mathbf{M}_\text{t}, \mathbf{M}_\text{sp}, \mathbf{M}_\text{spt}\}$ is the warping matrix cross camera coordinates and the suffix $t$, $sp$, $spt$ denotes temporal, spatial, and spatial-temporal correlations. $\mathbf{E}_i$ and $\mathbf{K}_i$ are extrinsic and intrinsic parameters of camera $i$.
Next, we use the photometric loss function \cite{wang2004image} to minimize the differences between the reference image and the reconstructed image,
\begin{equation}
    \mathcal{L} = \frac{\alpha}{2}\left[
   1-\operatorname{SSIM}\left(\mathbf{I}^l, \tilde{\mathbf{I}}^{t'}\right)\right]
   +
    (1-\alpha)\left\|\mathbf{I}^t-\tilde{\mathbf{I}}^{t'}\right\|_1 ,
\end{equation}
\noindent where the former term is to measure structure similarity and the latter is to punish identical differences. We adopt this loss for the aforementioned three types of contexts. Consequently, the total self-supervised loss can be formulated as:
\begin{equation}
    \mathcal{L}_{\text {CAST}}=\lambda_{\text {t }}\mathcal{L}_{\mathrm{t}}+\lambda_{\text {sp }} \mathcal{L}_{\text {sp }}+\lambda_{\text {spt}} \mathcal{L}_{\text {spt} }.
\end{equation}

%% file: exp.tex
\section{Experiments}
\subsection{Experimental Setup}
{\bfseries Dataset.}
Our experiments are conducted on the Occ3D-nuScenes \cite{tian2023occ3d} dataset, which comprises 700 scenes for training and 150 scenes for validation. Each frame contains LiDAR point cloud data of 32 lines and images captured by six surround-view cameras. 
Occ3D-nuScenes provides surround-view data captured by six cameras, making it highly suitable for our occupancy prediction task. Moreover, the dataset includes manually annotated occupancy ground truth, which facilitates supervised training.
The 3D occupancy ground-truths are provided at a resolution of $200 \times 200 \times 16$ voxels, encompassing a spatial range of $\left[\pm 40 \text{m}, \pm 40 \text{m}, -1 - 5.4 \text{m} \right]$. Each voxel is a cube with an edge of 0.4m. The dataset delineates 17 semantic categories, with an additional $free$ label denoting unoccupied space.

\noindent{\bfseries Evaluation Metrics.}
For the 3D occupancy prediction task, we compute the intersection over union (IoU) metric for scene completion and the mean IoU (mIoU) of all semantic classes for semantic occupancy predictions.

\begin{equation}
\begin{aligned}
\renewcommand{\arraystretch}{1.5}
\begin{array}{c}
\mathrm{IoU}=\frac{T P}{T P+F P+F N} \\
\mathrm{mIoU}=\frac{1}{K} \sum_{i=1}^{K} \frac{T P_{i}}{T P_{i}+F P_{i}+F N_{i}} \text{,}
\end{array}
\end{aligned}
\end{equation}
where $TP$, $FP$, and $FN$ denote the number of true-positive predictions, false-positive predictions, and false-negative predictions, respectively. $K$ denotes the category number.

\noindent{\bfseries Implementation Details.}
For \textbf{View Transformation}, we use Explicit-Implicit depth modeling to do 2D-3D view transformation. For EDM, we use explicit depth distribution to transfer the image feature to OCC feature 
$\mathbf{O}_e\in\mathbb{R}^{64\times200\times200\times16}$. For IDM, we first pre-define a learnable occupancy query $\mathbf{Q}\in\mathbb{R}^{64\times100\times100\times8}$ and conduct cross attention between query and image feature. Finally, we use trilinear interpolation to upsample the feature to raw resolution, obtain $\mathbf{O}_i\in\mathbb{R}^{64\times200\times200\times16}$. The occupancy feature is the concatenation of $\mathbf{O}_e$ and $\mathbf{O}_i$ in feature dimension $\mathbf{O}=\mathbf{O}_e\oplus\mathbf{O}_i$, $\mathbf{O}\in\mathbb{R}^{128\times200\times200\times16}$. The lightweight convolution layer adapted to compress and fuse the concatenated feature $\mathbf{O}^{\prime}\in\mathbb{R}^{128\times100\times100\times8}$.
For \textbf{Mask-based Encoder Decoder}, we use windowed attention and downsample convolution, marked as $\mathbb{F}_{enc}$, to produce multi-level feature $\mathbf{G}_i$. Concretely, $\mathbf{G}$ has three level, $\mathbf{G}_0\in\mathbb{R}^{192\times100\times100\times8}$, $\mathbf{G}_1\in\mathbb{R}^{192\times50\times50\times4}$ and $\mathbf{G}_2\in\mathbb{R}^{192\times25\times25\times2}$. Then, the multi scale deformable cross attention \textbf{3D-MSDA} is used to aggregate multi-level features. The mask-decoder $\mathbb{F}_{dec}$ use $\mathbf{G}_0,\mathbf{G}_1,\mathbf{G}_2$ to do predict the mask at resolution with $100\times100\times8$, then interpolate the logits to full resolution with $200\times200\times16$ to predict per-voxel semantic label.
For \textbf{Context-aware Self-training}, we use the occupancy feature of highest resolution $\mathbf{G}_0\in\mathbb{R}^{192\times100\times100\times8}$ to render surround view depth. In pretraining stage, the loss function has three components $\mathcal{L}_{pre}=\mathcal{L}_{ed}+\mathcal{L}_{rd}+\mathcal{L}_{CAST}$, where $\mathcal{L}_{ed}$ is depth supervision for explicit depth prediction in EDM module; $\mathcal{L}_{rd}$ is depth supervision for rendered depth; $\mathcal{L}_{CAST}$ is context-aware self-training loss. In training stage, the total loss is $\mathcal{L}_{train}=\mathcal{L}_{ed}+\mathcal{L}_{mask-cls}$, where $\mathcal{L}_{mask-cls}$ is occupancy mask prediction loss and $\mathcal{L}_{ed}$ is used, just like our baseline method. During the pretraining stage, the resolution of the rendered depth map is $180\times 320$. The number of sampled points on each ray is $S =152$. For context-aware self-training, we use six surround-view images and the corresponding rendered depth maps at time step $t$ and $t'=t-1$, respectively. We set $\alpha=0.85$ , $\lambda_\text{t}=1$, $\lambda_\text{sp} = 0.1$ and $\lambda_\text{spt} = 0.03$. Geometric pretraining is conducted in four epochs, followed by 24 epochs of supervised training. 
An AdamW \cite{loshchilov2017decoupled} optimizer with an initial learning rate of 1e-4 and weight decay of 0.01 is used.

\subsection{Main Results and the Discussion}
In Tab. \ref{tab: main results}, we present a comprehensive comparison of our method against leading methods on the Occ3d-nuScenes validation set. Furthermore, Tab. \ref{tab:per-class} shows a more detailed comparison for each semantic category. 
The experiments in Tab. \ref{tab: main results} and Tab. \ref{tab:per-class} involve the Long-Term Sequence Fusion Module and utilize two scales of image backbone: ResNet-50 and SwinTransformer-B.

\begin{table}[t]
\centering
\renewcommand\arraystretch{0.5}
\caption{Ablation study of each component under ResNet-50 image backbone. }
\begin{tabular}{cccc|c}
\toprule
\multicolumn{4}{c|}{Component} & Metric  \\ 
\midrule
EDM & IDM      & Mask-ED     & CAST   & mIoU(\%) \\
\midrule
 \CheckmarkBold  &         &         &        & 37.23        \\
    &     \CheckmarkBold     &         &        & 27.39       \\
\CheckmarkBold   & \CheckmarkBold        &         &         & 38.01         \\
 \CheckmarkBold & \CheckmarkBold        & \CheckmarkBold       &    &   40.25         \\
\CheckmarkBold & \CheckmarkBold        & \CheckmarkBold       & \CheckmarkBold     &   40.50 \\
\bottomrule
\end{tabular}
\setlength{\belowcaptionskip}{-0.3cm}

\label{tab: ablation}
\end{table}

\begin{table}[t]
\setlength{\belowcaptionskip}{-0.3cm}
    \caption{Ablation study on computational efficiency.}
    \centering
    \begin{tabular}{l|ccc}
    \hline
    method&Params(M)&Latency(s) & mIoU(\%)\\
    \hline
    BEVDet4D(R50) & 35&0.41&40.43\\
    \rowcolor{gray!20} Ours(R50)& 64&0.45&43.64\\
    FB-OCC(R50)& 68&0.30&42.14\\
    OccFormer(R50) &147& 0.32&37.4\\
    PanoOCC(R101) & 98&0.55&42.13\\
    BEVDet4D(SwinB)& 122&1.42&43.29\\
    \rowcolor{gray!20}Ours(SwinB)& 151&1.46&44.67\\
    \hline
    \end{tabular}
 \vspace{-2.5mm}
 \label{tab:latency}
\end{table}

\begin{table}[t]
\centering
\caption{Comparison of training phase that CAST used in.}
\begin{tabular}{l|cc}
\toprule
Training phase     & Params [M] & GFLOPs \\
\midrule
CAST in pretraining&64&472\\
CAST in training&64&540\\
\bottomrule
\end{tabular}

\label{tab:self-training}
\end{table}

\begin{figure}[t]
 \centering
 \includegraphics[width=\columnwidth]{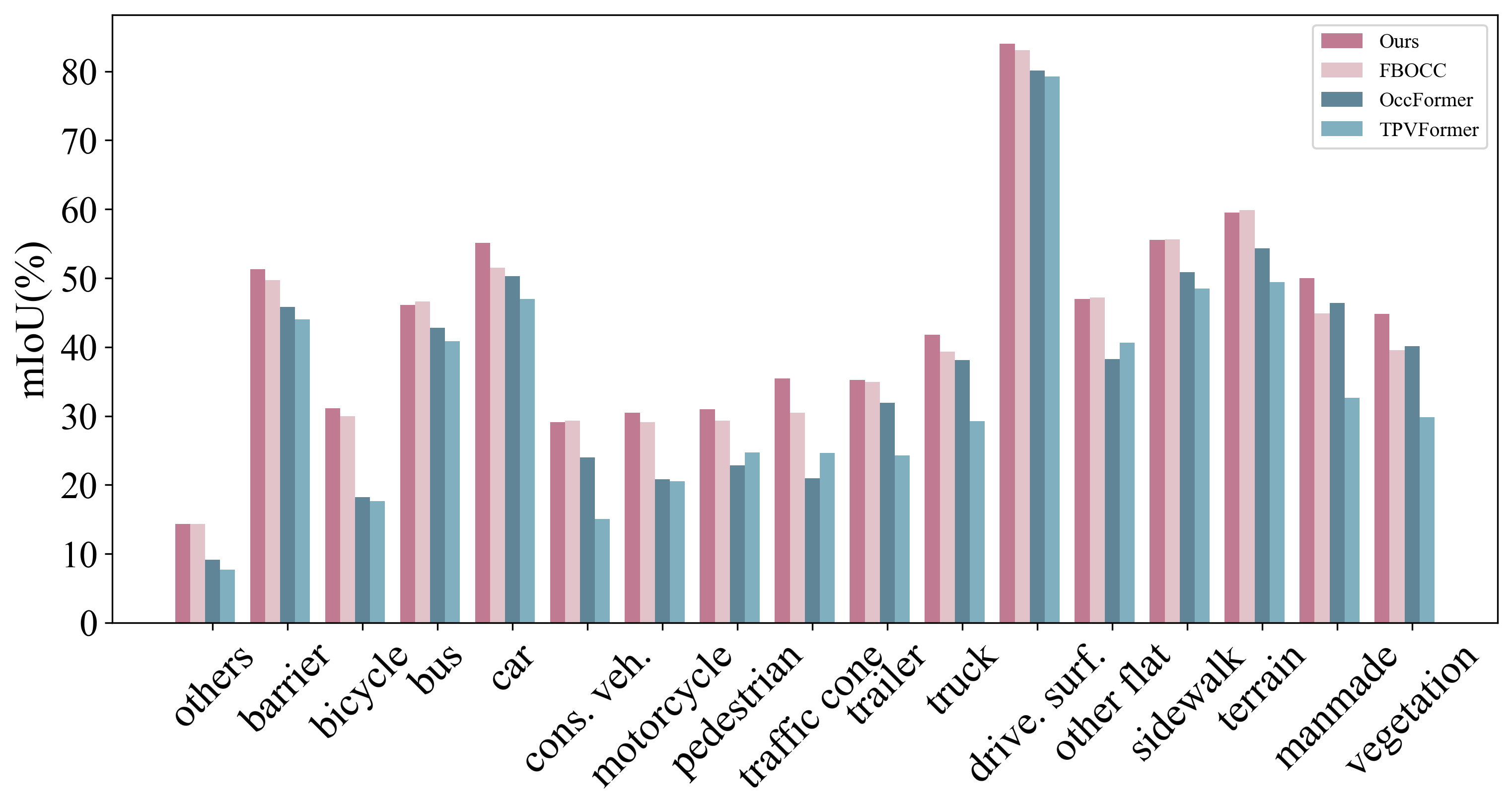} 
 \setlength{\abovecaptionskip}{-0.3cm}
 \setlength{\belowcaptionskip}{-0.4cm}
 \caption{
Per-class mIoU comparison with the previous SOTA methods. Our approach achieves better performance in most classes. The class-wise comparisons between
different methods are all under the image resolution of 256x704 and backbone network of ResNet-50 respectively.
 }
 \label{fig:sec4_per-class}
\end{figure}

\newcommand{\sq}[3]{\textcolor[RGB]{#1,#2,#3}{$\blacksquare$}}
\begin{table*}[htbp]
\caption{3D Occupancy prediction on Occ3d-nuScenes dataset. The performance of each semantic category compared with other sota methods.}
\centering
\resizebox{\textwidth}{!}{
\begin{tabular}{l|c|c|c|c|ccccccccccccccccc}
\toprule
Method&Mask&Backbone&IoU&mIoU&
\rotatebox{90}{\sq{0}{0}{0} others}&
\rotatebox{90}{\sq{70}{80}{90} barrier}&
\rotatebox{90}{\sq{220}{20}{60} bicycle}&
\rotatebox{90}{\sq{255}{127}{80} bus}&
\rotatebox{90}{\sq{255}{158}{0} car}&
\rotatebox{90}{\sq{233}{150}{70} cons. veh.}&
\rotatebox{90}{\sq{255}{61}{99} motorcycle}&
\rotatebox{90}{\sq{0}{0}{230} pedestrian}&
\rotatebox{90}{\sq{47}{79}{79} traffic cone}&
\rotatebox{90}{\sq{255}{140}{0} trailer}&
\rotatebox{90}{\sq{255}{99}{71} truck}&
\rotatebox{90}{\sq{0}{207}{191} drive. surf.}&
\rotatebox{90}{\sq{175}{0}{75} other flat}&
\rotatebox{90}{\sq{75}{0}{75} sidewalk}&
\rotatebox{90}{\sq{112}{180}{60} terrain}&
\rotatebox{90}{\sq{222}{184}{135} manmade}&
\rotatebox{90}{\sq{0}{175}{0} vegetation}\\
\midrule
Occformer&\CheckmarkBold&Res-50&70.1&37.4&
9.15&45.84&18.20&42.80&50.27&24.00&20.80&22.86&20.98& 31.94&38.13&80.13&38.24&50.83&54.3&46.41&40.15\\
BEVDet4D&\CheckmarkBold&Res-50&74.53&40.43&
10.72&48.72&19.6&43.61&52.82&27.88&22.8&24.58&24.45&35.31&40.61&83.25&41.65&55.01&58.07&51.4&46.79\\
FBOCC&\CheckmarkBold&Res-50&-&42.1&
14.30&49.71&30.0&46.62&51.54&29.3&29.13&29.35&30.48&34.97&39.36&83.07&47.16&55.62&59.88&44.89&39.58\\
\textbf{Ours}&\CheckmarkBold&Res-50&74.63&43.64&
14.29&51.27&31.11&46.13&55.09&29.12&30.46&30.99&35.47&35.2&41.82&84.0&47.0&55.52&59.5&50.03&44.82\\
\midrule
MonoScene&\XSolidBrush&Res-101&-&6.06&
1.75&7.23&4.26&4.93&9.38&5.67&3.98&3.01&5.90&4.45&7.17&14.91&6.32&7.92&7.43&1.01&7.65\\

BEVDet&\XSolidBrush&Res-101&-&11.73
&2.09&15.29&0.0&4.18&12.97&1.35&0.0&0.43&0.13&6.59&6.66&52.72&19.04&26.45&21.78&14.51&15.26\\
BEVFormer&\XSolidBrush&Res-101&-&23.67&
5.03&38.79&9.98&34.41&41.09&13.24&16.50&18.15&17.83&18.66&27.70&48.95&27.73&29.08&25.38&15.41&14.46
\\
BEVStereo&\XSolidBrush&Res-101&-&24.51
&5.73&38.41&7.88&38.70&41.20&17.56&17.33&14.69&10.31&16.84&29.62&54.08&28.92&32.68&26.54&18.74&17.49\\
TPVFormer&\XSolidBrush&Res-101&-&28.34
&6.67&39.20&14.24&41.54&46.98&19.21&22.64&17.87&14.54&30.20&35.51&56.18&33.65&35.69&31.61&19.97&16.12\\
CTF-Occ&\XSolidBrush&Res-101&-&28.53
&8.09&39.33&20.56&38.29&42.24&16.93&24.52&22.72&21.05&22.98&31.11&53.33&33.84&37.98&33.23&20.79&18.00\\
VoxFormer&\CheckmarkBold&Res-101&-&40.7
&-&-&-&-&-&-&-&-&-&-&-&-&-&-&-&-&-\\

PanoOcc&\CheckmarkBold&Res-101&-&42.13&
11.67&50.48&29.64&49.44&55.52&23.29&33.26&30.55&30.99&34.43&42.57&83.31&44.23&54.40&56.04&45.94&40.40\\
\midrule

RenderOcc&\XSolidBrush&Swin-B&-&26.11
&4.84&31.72&10.72&27.67&26.45&13.87&18.2&17.67&17.84&21.19&23.25&63.20&36.42&46.21&44.26&19.58&20.72\\
SurroundOcc&\CheckmarkBold&Intern-B&-&40.7
&-&-&-&-&-&-&-&-&-&-&-&-&-&-&-&-&-\\
\textbf{Ours}&\CheckmarkBold&Swin-B&72.64&44.67&
14.02&51.4&33.08&52.08&56.72&30.04&33.54&32.34&35.83&39.34&44.18&83.49&46.77&55.72&58.94&48.85&43.0\\
\bottomrule
\end{tabular}}
 \setlength{\belowcaptionskip}{-0.5cm}

\label{tab:per-class}
\end{table*}

\begin{figure*}[!htbp]
 \centering
 \includegraphics[width=1.0\textwidth]{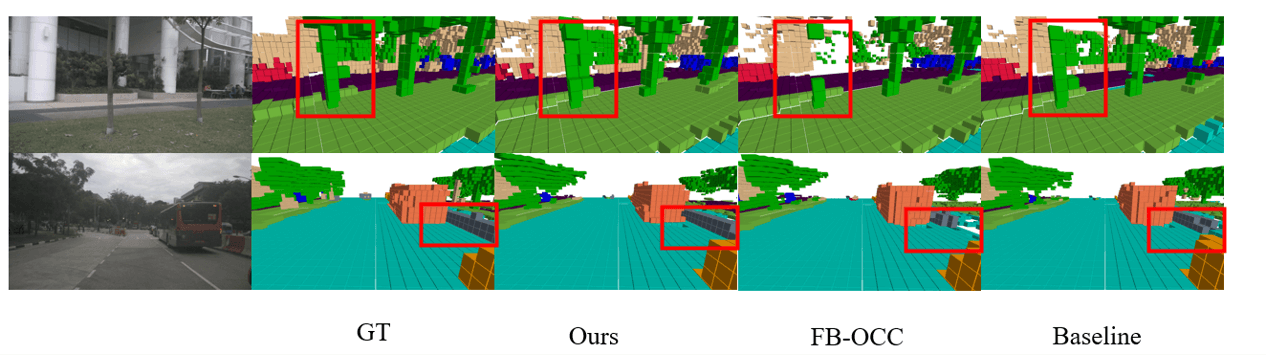} 
 \setlength{\abovecaptionskip}{-0.3cm}
 \setlength{\belowcaptionskip}{-0.5cm}
 \caption{
Qualitative visualization results for occupancy prediction in surround camera view. Better viewed when zoomed in.
 }
 \label{fig:sec4_camvis}
\end{figure*}

\noindent{\bfseries Results on ResNet-50.}
 With a ResNet-50 backbone, our approach surpasses the FB-OCC by 1.50\% mIoU and outperforms the BEVDet4D baseline by 3.21\% mIoU. Notably, our method, at a resolution of $256 \times 704$, not only exceeds BEVDet4D at $384 \times 704$ resolution using the same ResNet-50 backbone but also outstrips models like VoxFormer and PanoOcc with heavier backbones ResNet-101 and higher resolutions like $900 \times 1600$. Additionally, our ResNet-50-based method outperforms SurroundOcc with InternImage-B at $900 \times 1600$ resolution and slightly edges out the BEVDet4D baseline with SwinTransformer-B at $512 \times 1408$, underscoring its robustness and efficient generalizability.

\noindent {\bfseries Results on Larger Backbone.}
When employing larger backbones like SwinTransformer-B, our method retains SOTA performance, improving the BEVDet4D and SurroundOcc by 1.58\% mIoU. Despite a smaller improvement margin compared with ResNet-50 backbones, this could be due to their finetuning, which might lead larger models to overfit on the Occ3D-nuScenes dataset, thus diminishing the benefits of our GEOcc Network.

\noindent {\bfseries Comparing with UniOcc.}
Compared to UniOcc, which uses LiDAR depth and \textit{lidarseg} labels projected onto 2D to supervise depth and semantic maps, our GEOcc does not require additional LiDAR annotations for self-supervising rendered pure depth maps. Our method achieves a 3.94\% mIoU increase over UniOcc with a ResNet-50 backbone and is only marginally lower by 0.53\% mIoU when using a smaller resolution and lighter backbones, proving the competitiveness of our self-supervised rendering module against fully-supervised approaches.

\subsection{Ablation Studies \& Qualitative Visualization}
{\bfseries Effectiveness of Each Component.}
Since the training time is very long, we use the smallest image resolution of 256x704 and the smallest image backbone ResNet-50 in the ablation experiments, and also remove the long-term sequence fusion module to reduce the training time.
The ablation study, presented in Tab. \ref{tab: ablation}, shows our baseline method achieving a 37.23\% mIoU using EDM alone. Replacing EDM with IDM led to an approximate 10\% drop in mIoU, while their combination yielded a 0.8\% mIoU improvement. 
Our method combines explicit and implicit depth modeling, achieving superior feature robustness compared to these approaches. Additionally, for the classification head, our GEOcc employs a mask-based prediction head, enabling more precise predictions.
Incorporating the Mask-based transformer Encoder-Decoder (Masked-ED) structure further enhanced the model's feature perception, resulting in a 1.80\% mIoU increase. The implementation of Context-Aware Self-Training (CAST) loss improved performance by an additional 0 .25\% mIoU, optimizing network initialization without increasing parameter count.
Our CAST module provides the network with rich geometric priors during the pretraining phase, enabling the model to gain an initial perception of spatial occupancy. This allows the model to achieve rapid convergence during the subsequent supervised training phase.
Together, these enhancements contributed to a 3.27\% mIoU improvement under the ResNet-50 backbone. After the ablation experiments, we re-add the temporal fusion module, thus the final performance is consistent with that 43.64 reported in Tab. \ref{tab: main results}.

\begin{figure}[t]
 \centering
 \includegraphics[width=0.9\columnwidth]{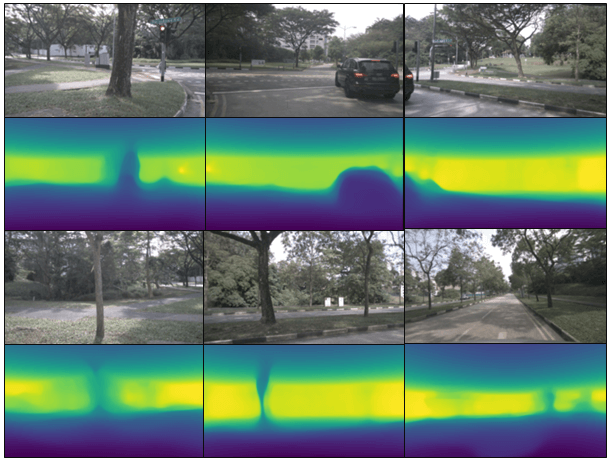} 
 \setlength{\belowcaptionskip}{-0.4cm}
 \caption{
    Visualization of depth maps in the surround view after geometric pretraining. Brighter colored areas indicate deeper depth, and vice versa.
 }
 \label{fig:sec4_depth}
\end{figure}

\noindent{\bfseries The Effectiveness of Neural Rendering.}
In the geometric pretraining phase, our network gains the ability to perceive spatial geometry without 3D occupancy label supervision.
As can be seen from the visualization results of Fig. \ref{fig:sec4_depth}, the network rendered a reasonable depth estimation of the surround-view images, where many salient landmarks, such as the road, vehicles, and vegetation in the near field can be reconstructed well. Geometric learning possesses enhanced prior knowledge of salient objects and their positional embedding in the 3D field, which can boost a faster convergence rate for semantic learning afterward.

\noindent{\bfseries Ablation on Scales of Model Parameters.}
Table \ref{tab:latency} shows the comparison between our method and other methods in terms of the number of model parameters and the amount of computation. For a fair comparison, all experiments were performed on a single NVIDIA RTX A6000 GPU, and the inference batch size is set to 1. The latency(s) is used to record computational time.

We would like to point out that as all current Occupancy-based methods \cite{huang2022bevdet4d,pan2023uniocc,ma2023cotr}, which cannot achieve real-time performance, our method is not designed for online driving, although it shows great potential.
However, the greater significance of GEOcc for autonomous driving lies in its ability to construct a high-precision offline ground truth system. Our method achieves the highest accuracy, making it ideal for simulation training and also service for autonomous driving.

Examining Tab. \ref{tab:self-training}, a comparative analysis is conducted regarding the integration of our CAST into distinct training phases. It is evident that employing our CAST during the training phase significantly amplifies FLOPS, which inspires us to use a pretraining strategy.

\begin{figure}[t]
 \centering
 \includegraphics[width=0.51\textwidth]{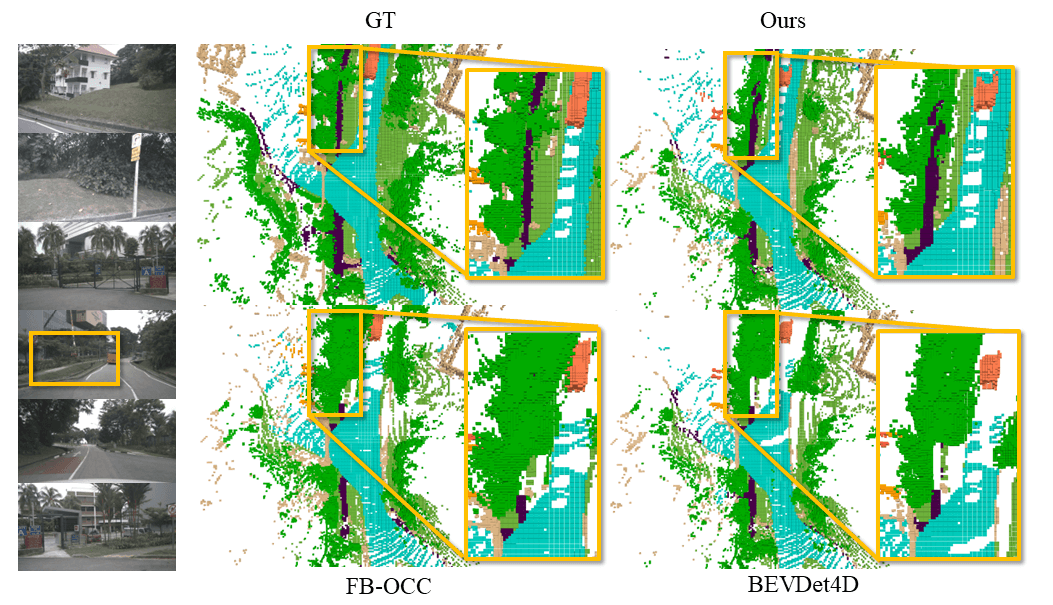} 
 \setlength{\abovecaptionskip}{-0.1cm}
 \setlength{\belowcaptionskip}{-0.4cm}
 \caption{
Qualitative visualization results for occupancy prediction in BEV view.
 }
 \label{fig:sec4_bevvis}
\end{figure}

\noindent{\bfseries Qualitative Visualization Results}
 Our method is compared with two high-performance benchmarks, BEVDet4D and FB-OCC, as illustrated in Figs. \ref{fig:sec4_camvis} and \ref{fig:sec4_bevvis}. 
From the first row of Fig. \ref{fig:sec4_camvis}, it can be seen that our method predicts more complete results for roadside trees, while the other methods show omissions. The second row demonstrates that our method accurately identifies roadside water barriers, whereas the other methods also exhibit missing predictions. These qualitative comparisons highlight the advantages of our method over the others.
 In terms of geometric reconstruction, GEOcc provides superior continuity for slender and low-lying objects, such as tree trunks and fences, as shown by the green and gray grids, respectively, in Fig. \ref{fig:sec4_camvis}.

\begin{figure}[t]
 \centering
 \includegraphics[width=0.5\textwidth]{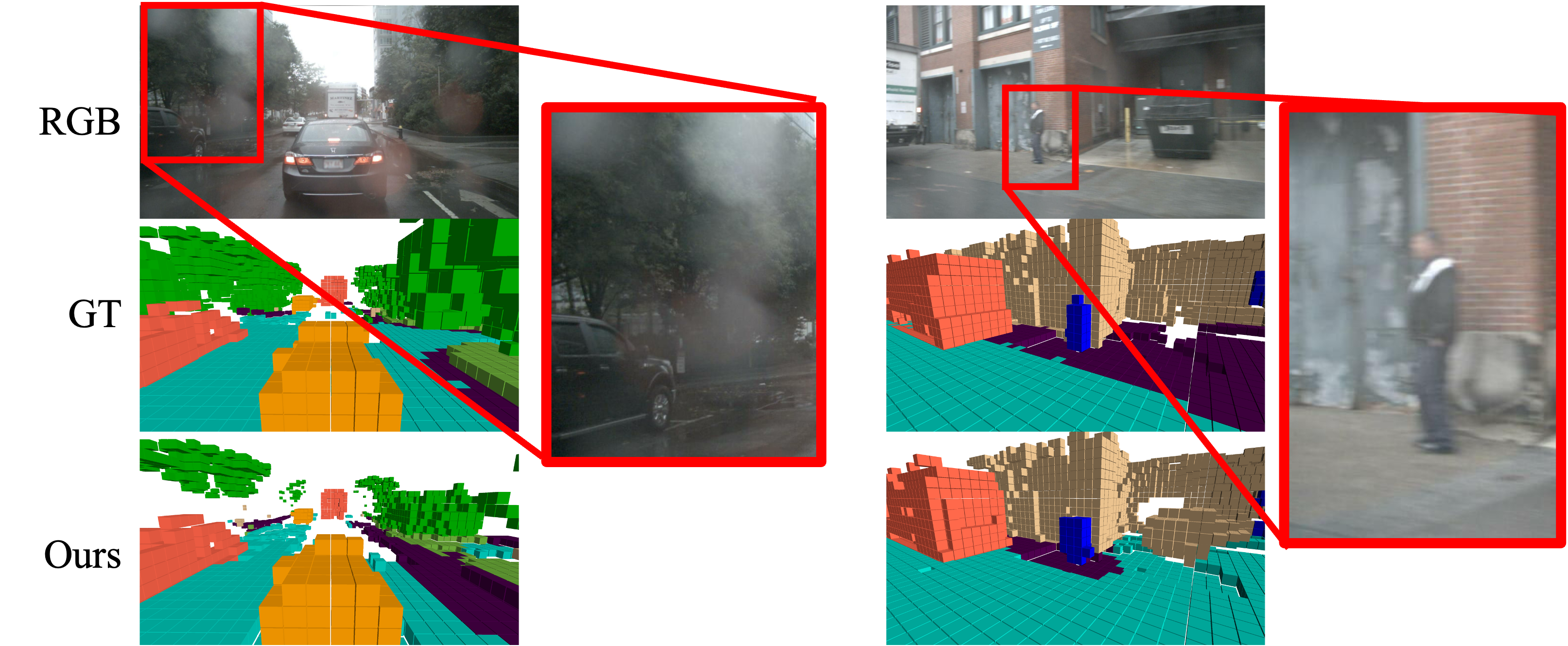} 
 \setlength{\abovecaptionskip}{-0.1cm}
 \setlength{\belowcaptionskip}{-0.4cm}
 \caption{Visualization analysis of model performance under low-quality imaging conditions. The left shows a rainy scene. The right illustrates motion blur caused by road bumps.}
 \label{fig:rain&blur}
\end{figure}

\begin{figure}[t]
 \centering
 \hspace{-1cm}
 \includegraphics[width=0.5\textwidth]{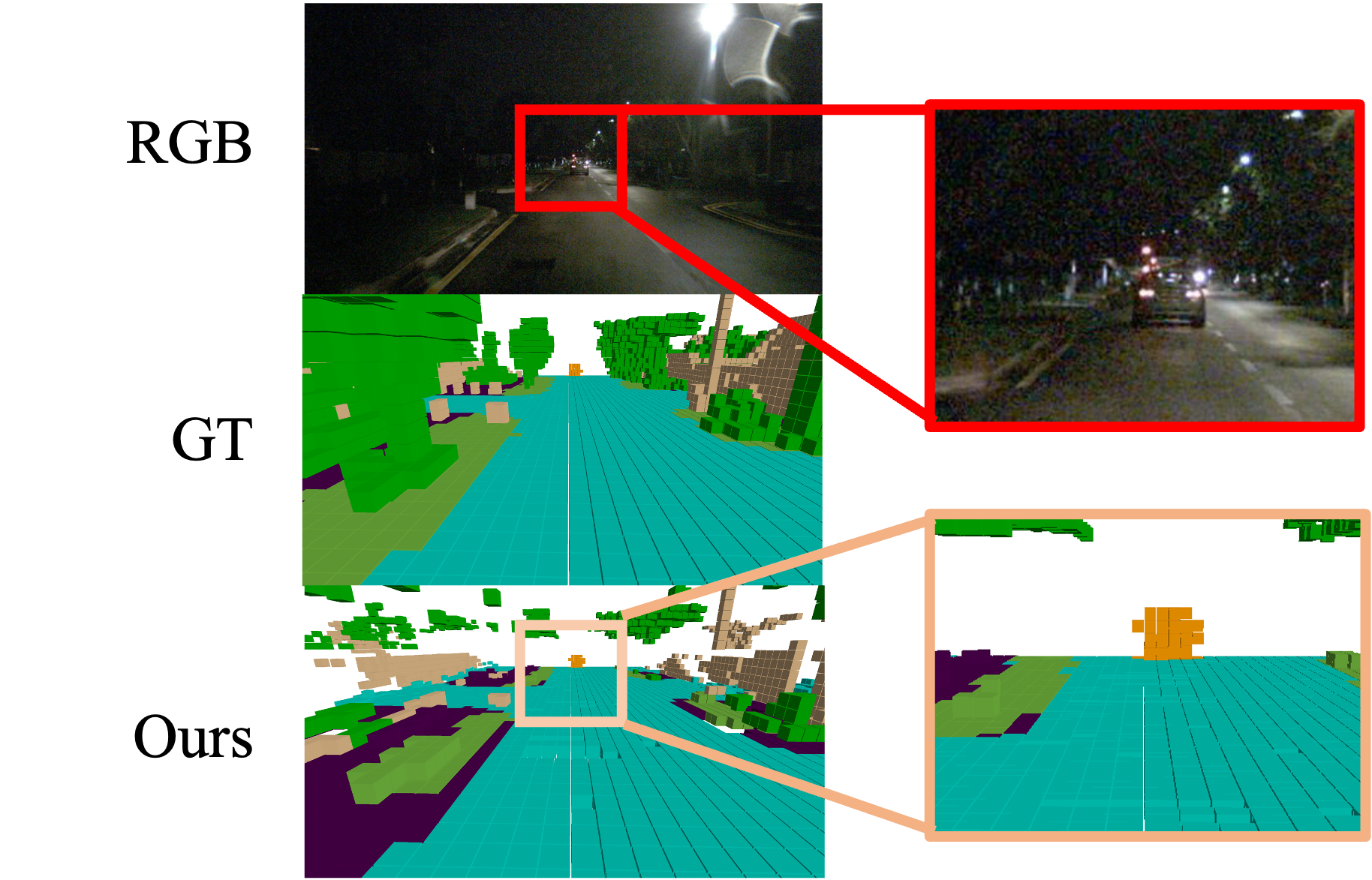} 
 \setlength{\abovecaptionskip}{-0.1cm}
 \setlength{\belowcaptionskip}{-0.4cm}
 \caption{Visualization of occupancy prediction results under low light conditions.}
 \label{fig:low-light}
\end{figure}

\noindent
{\bfseries Visualization Analysis of extreme weather condition and low lighting scene}
We conduct further experimental analysis on the specific environmental conditions involved in the dataset. 
On the left of Fig. \ref{fig:rain&blur}, raindrops adhering to the camera lens obscure imaging of vehicles and trees. On the right, rapid vehicle speed or road bumps result in motion blur, affecting the imaging of pedestrians along the roadside.
Despite the disturbances caused by weather conditions and traffic situations, the visualization results show that our method achieves performance very close to the ground truth, with precise recognition of key objects. This demonstrates the robustness of our approach under challenging weather and traffic conditions.
From the Fig. \ref{fig:low-light}, we can observe that under low-light conditions, a camera positioned directly in front of the vehicle captures an image of a car located less than 40 meters ahead. The spatial occupancy grid predicted by our method clearly highlights the presence of the car, even in a nighttime setting with insufficient lighting and a noisy imaging environment.
 
%% file: conclusion.tex
\section{The Conclusion and Limitation}
This work introduces GEOcc, a Geometrically Enhanced 3D occupancy network, using implicit-explicit depth fusion and contextual self-supervision. Integrating implicit and explicit depth modeling, GEOcc provides a robust 3D occupancy representation. An advanced mask-based encoder-decoder ensures finer semantic prediction, while context-aware self-training enriches geometric priors. GEOcc achieves state-of-the-art results on Occ3d-nuScenes dataset.
While, the primary limitation lies in the time bottleneck associated with the pixel-wise volume rendering component in the self-training phase. In our future work, we will focus on the optimization of volume rendering, as well as pre-training using large-scale unlabeled datasets.
We hope this work inspires further research in vision-based occupancy prediction.
